\documentclass[11pt,letterpaper]{article}
\usepackage[T1]{fontenc}
\usepackage{mathptmx}
\usepackage[scaled=0.92]{helvet}
\usepackage[letterpaper,margin=1in]{geometry}
\usepackage{booktabs}
\usepackage{array}
\usepackage{graphicx}
\usepackage{float}
\usepackage[round,authoryear]{natbib}
\usepackage{caption}
\usepackage{url}
\usepackage[hidelinks]{hyperref}
\newcommand{\doi}[1]{\href{https://doi.org/#1}{\nolinkurl{#1}}}
\newcommand{\sysname}[1]{\texttt{#1}}

\newcommand{\idn}[1]{#1}


\title{Calibrated Evidence Inference With a Small, Permissively Licensed Open Model}
\author{Johann Emmanuel Li\\[2pt]
\small ORCID iD \href{https://orcid.org/0009-0009-8726-460X}{\nolinkurl{0009-0009-8726-460X}}}
\date{October 2026}

\begin{document}
\maketitle

\begin{abstract}
\noindent Systematic reviewers who use a model to read randomized trials need each answer and a measure of how far to trust it. If the model is calibrated, so that its stated probabilities match how often it is right, a reviewer can accept its confident answers and check the rest by hand. This paper fine-tunes Qwen3-4B-Base with low-rank adapters and a classification head for evidence inference \citep{lehman2019,deyoung2020}: given a trial's article and a prompt that names an intervention, a comparator and an outcome, classify the result as a significant increase, a significant decrease or no significant difference. The articles of Evidence Inference 2.0 come from the open-access subset of PubMed Central, which includes articles under non-commercial and no-derivatives licenses, so the released model was fine-tuned only on the 1,431 of 2,657 training articles under CC BY, CC0 or the public domain; it is released under the Apache License 2.0 and runs on a laptop. On the test split, its expected calibration error is 0.017 and its log loss is lower than that of Gemma 4 31B-it given the same input zero-shot, meeting criteria registered on the Open Science Framework before testing. Its macro-F1, 0.925, is above every published result on that split that I found, though those systems used other inputs, scoring rules and prompt filters, and the test labels are public. Accepting its 80 percent most confident answers leaves an error rate of 2.6 percent among them. Context ablations show that the answers depend on the article, but the title alone carries some signal, which an informative title or recall of the trial could explain.
\end{abstract}

\section{Introduction}
\label{sec:intro}

A systematic reviewer, or a service that keeps an evidence summary up to date, needs two things from a model that reads randomized trials: the answer, and how far to trust it. If an answer's stated probability matches how often answers with that probability are right, the reviewer can accept the confident answers and check only the uncertain ones by hand. That requires calibrated probabilities, and it requires that the calibration be measured on data the model was not tuned on.

Small models matter for this because they are cheap to run and can run locally: a small model costs less per question than a large one (Appendix~\ref{app:cost}), and a model that runs on a reviewer's own computer can read documents that may not leave an institution. Yet small language models in clinical medicine seldom report calibration: a systematic review of clinical language models of up to 4 billion parameters found 11 eligible studies, and none reported a scalar calibration error \citep{gorenshtein2026}.

This paper reports the evidence inference model of Fiorillo v0.5, an open release of task-specific models for closed-set medical classification, each of which gives a calibrated probability for every label. The task is that of Evidence Inference 2.0 \citep[EI;][]{lehman2019,deyoung2020}: given a randomized trial's article and a prompt that names an intervention, a comparator and an outcome, did the outcome significantly increase, significantly decrease, or show no significant difference? The release also holds a PubMedQA model, from which the EI model is initialized, and a hospital-readmission model trained on tabular data; Appendix~\ref{app:other} reports both. Two EI models appear throughout: the released model, fine-tuned only on permissively licensed articles, and a full-data model, fine-tuned on all of EI's training articles, which was the subject of the original registration and stays private. The paper makes four contributions:
\begin{itemize}\setlength{\itemsep}{1pt}
\item an open 4-billion-parameter EI model, fine-tuned only on training articles under CC BY, CC0 or the public domain and released under the Apache License 2.0, which gives calibrated probabilities for the three labels;
\item a pre-registered evaluation on EI's test split, the split's second pre-registered use, in which the released model meets all four criteria fixed before any of its test predictions: a limit on its calibration error, a lower log loss than a class-prior baseline and than Gemma 4 31B-it given the same input zero-shot, and a macro-F1 not below Gemma 4 31B-it's; its macro-F1 is above every published result on that split that I found, though not in a controlled comparison;
\item a measurement of what license filtering costs: the same training configuration was less accurate when fine-tuned only on the permissively licensed articles;
\item context ablations (no article, the title only) and a counterfactual test that swaps the intervention and the comparator, which ask whether the answers depend on the article, and a selective-prediction analysis of how many answers a reviewer could accept without checking them.
\end{itemize}

\section{Related work}
\label{sec:related}

\paragraph{Evidence inference.} \citet{lehman2019} introduced the task and the dataset, and \citet{deyoung2020} released Evidence Inference 2.0, with more prompts, and its splits; a classifier given the prompt alone, without the article, reached a macro-F1 of 0.511 in their Table 1. Published results on EI's test split were scored and counted in different ways: \citet{deyoung2020} report a full-text pipeline and a classifier given the annotated evidence spans, and \citet{buchmann2024structure} report long-document transformers scored leniently, where a prediction that matches any of a prompt's accepted labels counts as right. Table~\ref{tab:published} sets these values beside this paper's results. \citet{buchmann2024lab} include EI in a benchmark of long-document assistants and report macro-F1 values of 0.86 and 0.87 for models answering from the whole article or from retrieved segments (single runs, scored by their own rules, one of them on 100 test instances); they were not part of the pre-registered comparison and are not in the table. EI is also one of the datasets of the ERASER benchmark \citep{deyoung2020eraser}, where a model must also return the evidence that supports its answer; the model here returns no evidence span. Later work recasts EI's annotations: \citet{wadhwa2023} train large language models on data derived from EI to extract interventions, comparators, outcomes and findings jointly from abstracts, and SciRIFF \citep{wadden2025} includes EI among its held-out tasks as the extraction of intervention, comparator, outcome, effect and evidence tuples from abstracts. Those results are for other task definitions and are not in the table. Trialstreamer \citep{marshall2020} keeps a living, automatically updated database of randomized trial reports, with their populations, interventions and outcomes extracted by machine learning.

\paragraph{PubMedQA and BioGPT.} PubMedQA \citep{jin2019} asks yes, no or maybe questions about research abstracts; BioGPT \citep{luo2022}, a biomedical language model, has a version fine-tuned on it, which this paper runs as a reference (Appendix~\ref{app:other}).

\paragraph{Calibration and selective prediction.} The Brier score \citep{brier1950}, the expected calibration error \citep{naeini2015}, temperature scaling \citep{guo2017} and Dirichlet calibration \citep{kull2019} are the measures and methods used here. Answering only the inputs on which a model is most confident is selective prediction \citep{geifman2017}; sending the rest to a larger model is a cascade with deferral. Both are evaluated here with risk-coverage measures, including the area under the generalized risk-coverage curve that \citet{traub2024} recommend.

\paragraph{Contamination and pre-registration.} A model may have seen a benchmark's test items or labels during pretraining \citep{sainz2023}. \citet{windisch2026} found that frontier models classified oncology trials from their titles alone, which bears on the title-only ablation in Section~\ref{sec:reading}. Pre-registering an analysis before the data are seen has been proposed for natural language processing research by \citet{vanmiltenburg2021}.

\section{Data and licenses}
\label{sec:data}

\paragraph{Evidence Inference 2.0.} I used EI at a fixed commit of its repository; its prompts and annotations are under the MIT License. After the exclusions the registration states, the splits hold: training, 2,657 articles and 10,046 prompts; validation, 340 articles and 1,237 prompts; test, 333 articles and 1,218 prompts. The exclusions are the prompts with no accepted annotation (210, 21 and 28 in the three splits), the 70 training prompts that EI's annotations list as Incorrect, the 6 test prompts of an article whose text is empty, and 29 prompts in no split. A prompt's label is the most common label among its accepted annotations; a tie goes to the smallest of EI's label codes ($-$1 for a significant decrease, 0 for no significant difference, 1 for a significant increase), which is what SciPy's mode returns and what EI's own code uses.

\paragraph{License filtering.} EI's articles come from the open-access subset of PubMed Central (PMC), whose articles keep their own licenses: PMC groups them as allowing commercial use (CC0, CC BY, CC BY-SA and CC BY-ND), as non-commercial use only (the CC BY-NC licenses), or as other (a custom license, or none that a machine can read). A script read each article's own license statement as PMC serves it, through the National Center for Biotechnology Information's E-utilities, and classified it by fixed rules, which a separate review checked. An article is permissively licensed when its statement is the Creative Commons Attribution license (CC BY, any version, by link or in words), the Creative Commons public-domain dedication (CC0) or the public domain for the article itself, and no part of the statement restricts commercial use, derivatives or the license of what is made from it; where the parts disagree, the more restrictive part decides, a statement that names a publisher's own terms is not permissive even beside Creative Commons words, and CC BY-SA (Attribution-ShareAlike) is not permissive. The rule is thus stricter than PMC's commercial-use group, which includes the share-alike and no-derivatives licenses. By this rule, 1,431 of the 2,657 training articles were permissively licensed (5,279 of 10,046 prompts), 179 of the 340 validation articles (654 of 1,237 prompts) and 170 of the 333 test articles (613 of 1,218 prompts). The other 1,226 training articles are under non-commercial, no-derivatives, share-alike, verbatim-only or publishers' own terms, or state no license. Whether model weights trained on such articles may be released under an open license is, as far as I could find, legally unsettled, so the full-data model, which was fine-tuned on all of them, stays a private research model, and the released model's EI fine-tuning used only permissively licensed articles.

\paragraph{Two licensing caveats.} First, the passage selection (Section~\ref{sec:model}) ranks passages by Okapi BM25, a word-matching score \citep{robertson2009}, whose word statistics were counted over the passages of all 2,657 training articles, permissively licensed or not. They hold single words and their counts but no sentence of any article. I kept them so that every input, the released model's and Gemma 4 31B-it's, is cut by the same rule, and they ship with the release so that a new article can be cut the same way. Second, the released model was initialized from the weights of the PubMedQA model (Appendix~\ref{app:other}), whose training abstracts' licenses were not checked one by one.

\paragraph{Models used.} Qwen3-4B-Base \citep{qwen3}, the base model, and Gemma 4 31B-it \citep{gemma4}, the zero-shot reference, are both under the Apache License 2.0, as their model cards give it.

\section{The model}
\label{sec:model}

\paragraph{Architecture.} The EI model is Qwen3-4B-Base at a fixed revision with low-rank adaptation \citep[LoRA;][]{hu2022} adapters on every linear layer of the decoder (rank 16, alpha 32, dropout 0.05), a linear projection and a classification head: two transformer layers and a scorer that gives one logit for each class, with a softmax over the three classes. The base model's own weights stay fixed.

\paragraph{Training.} I initialized the model from the weights of the PubMedQA model (intermediate-task transfer) and trained it for two epochs, with inputs of 6,144 tokens and a learning rate of 0.00005 for the adapters, on the 5,279 permissively licensed training prompts. Each input holds the intervention, the comparator and the outcome, then the article's title and abstract, then passages chosen by BM25 against the prompt's words, results-section passages first, while they fit; the kept passages follow in their article order. The released model is a uniform model soup \citep{wortsman2022} of the chosen configuration's three runs: each layer's three LoRA updates are joined into one adapter of rank 48 (alpha 96) by concatenating their A matrices and their B matrices divided by three, so that its update equals the exact average of the three, and the projection's and head's weights are averaged, so one forward pass answers a question.

\paragraph{Model selection.} I fixed the selection rule before training. Four candidate changes to the baseline configuration (the full-data model's configuration, fine-tuned on the permissively licensed articles) were each trained twice: swap augmentation, which adds a copy of each training prompt with the intervention and comparator swapped; inputs of 6,144 tokens instead of 4,096; and two higher learning rates. Each was scored on the 654 permissively licensed validation prompts by the out-of-fold log loss of its calibrated probabilities (5 folds, 20 repeats). A candidate passed if its mean score was below the baseline's by more than a margin set from the baseline's own spread across three seeds, 0.0110, and each of its runs scored below the baseline run with the same seed. Only the longer input passed, and it was chosen; a third run completed it. The soup was kept because its score was not above that of the three runs' prediction ensemble (their probabilities averaged) by more than the margin. Swap test-time augmentation, which averages each answer with the answer to the swapped prompt, was not kept because it scored worse (Appendix~\ref{app:runs}).

\paragraph{Calibration.} I chose the calibration method among four by out-of-fold log loss on the 654 permissively licensed validation prompts: temperature scaling, 0.1781; temperature scaling with a bias for each class, 0.1808; vector scaling, 0.1836; and Dirichlet calibration with off-diagonal regularization, 0.1829. I fitted the chosen method, temperature scaling, on all 654 prompts: the temperature is 1.340.

\paragraph{No deferral.} The released model answers every question itself. In the full-data model's evaluation, deferring its least confident EI answers to Gemma 4 31B-it gave an estimated combined accuracy below the full-data model's alone at every deferral rate tested (no interval exists for that difference), so I fixed the released model's deferral rate at zero before training it.

\section{Pre-registered evaluation}
\label{sec:tests}

\paragraph{Registration.} I registered the evaluation of the full-data model (named Fiorillo v0.4 in the registration) on OSF \citep{osfreg} on 1 October 2026 at 00:36 UTC (Coordinated Universal Time), before any model had made a prediction on EI's test split, under an embargo. It fixed four criteria for that model, an average of three runs fine-tuned on all of EI's training articles, against a class-prior baseline and Gemma 4 31B-it, with the model and Gemma 4 31B-it each calibrated on validation by the same method. Gemma 4 31B-it is used zero-shot, with thinking off, and each answer's probability is read from its next-token probabilities for the option letters.

\paragraph{Amendment.} On 1 October 2026 at 15:31 UTC, before any test prediction of the released model existed, I filed an amendment to the registration. It keeps the four criteria, with the released model in the full-data model's place, and fixes ten comparisons, which it codes K1 to K10 (Table~\ref{tab:analyses} maps each to this paper's sections). The release candidate, with every calibration, the limit of criterion 1 below and the hash of each file they come from, was frozen on 2 October 2026 at 13:40 UTC, and the test predictions started after it.

\paragraph{The four criteria.} Each is computed over all 1,218 test prompts.
\begin{enumerate}\setlength{\itemsep}{1pt}
\item The expected calibration error (ECE; top-label, with 15 equal-width bins on the top answer's probability) is at most the limit: the larger of 0.05 and the 95th percentile of the ECE that a perfectly calibrated model shows by chance on 1,218 prompts, from 5,000 simulated test sets. For the released model that percentile was 0.0184, so the limit was 0.05.
\item The log loss is below the class-prior baseline's, and the 95 percent interval of the difference lies wholly below zero. The class-prior baseline gives every prompt the label distribution of the whole training split.
\item The log loss is below that of Gemma 4 31B-it given the same selected text, and the interval of the difference lies wholly below zero. For the released model, Gemma 4 31B-it read the same 6,144-token inputs.
\item The macro-F1 is not below Gemma 4 31B-it's point estimate.
\end{enumerate}
Intervals come from a paired bootstrap over articles (prompts from one article are not independent), with percentile intervals; criteria 2 and 3 use 10,000 article resamples, and every other EI interval 1,000. A difference whose interval includes zero is reported as ``no difference was detected''. Success requires all four criteria (an intersection-union test), so no correction for multiple comparisons applies to it. The registration's power analysis gave the full-data model 80 percent power for criterion 3 at a true log-loss improvement of 0.0787, and the amendment, before any test prediction of the released model, gave the released model the same power at 0.0718 (both from a normal approximation).

\paragraph{Primary and secondary analyses.} The four criteria are the primary analysis and the only confirmatory one. This paper calls every other pre-registered analysis secondary and reports it as an estimate with its 95 percent interval where one was computed; Table~\ref{tab:analyses} gives each one's code and the label the registration gave it. Analyses that were not pre-registered are marked exploratory where they appear. The registration states that its author had seen the data but had not run the proposed analyses. The released model's evaluation is the second use of EI's test split. The split was first used for the full-data model's criteria, its secondary analyses and its exploratory checks, and for its context ablations and swap test, which the amendment pre-registered; while the test examples were built, the counts of their labels were printed and read, and while the title-only input's rule was corrected, the test articles' title lines were read, but no label. Three of the released model's design choices drew on the full-data model's test results: no deferral, no larger base model, and the starting effect size of the power analysis; the amendment disclosed each. The released model's training configuration and calibration, and the release's readmission ensemble, were chosen on validation only. Appendix~\ref{app:deviations} lists every deviation from the registration and the amendment.

\section{Results}
\label{sec:results}

\paragraph{The full-data model.} The full-data model met all four criteria as registered (Table~\ref{tab:bar}): ECE 0.0119; log loss minus the class-prior baseline's $-$0.8773 ($-$0.9225 to $-$0.8281); minus Gemma 4 31B-it's $-$0.2084 ($-$0.2720 to $-$0.1505); macro-F1 0.9399 against 0.8626. Of its other pre-registered secondary analyses, this paper gives the comparison with published results (Table~\ref{tab:published}) and, in brief, the deferral analysis (Sections~\ref{sec:model} and~\ref{sec:handoff}), not its single runs, PubMedQA, readmission, sensitivity analyses or zero-probability check.

\paragraph{The released model.} The released model met all four criteria (Table~\ref{tab:bar}): ECE 0.0168 against the limit 0.05; log loss minus the class-prior baseline's $-$0.8603 ($-$0.9078 to $-$0.8104); log loss minus Gemma 4 31B-it's $-$0.1829 ($-$0.2598 to $-$0.1164); and macro-F1 0.9248 against 0.8668. Without the 6 test prompts that EI's annotations flag, on 1,212 prompts, the two log-loss differences were $-$0.8600 ($-$0.9087 to $-$0.8109) and $-$0.1848 ($-$0.2606 to $-$0.1164) (a pre-registered sensitivity analysis). Of the 194,637 predictions of every model, split and task, none gave the true label a probability that rounds to 0 in the results files.

\begin{table}[t]
\centering\small
\caption{The four pre-registered criteria on Evidence Inference 2.0's test split (1,218 prompts in 333 articles), for the full-data model as registered and for the released model as the amendment fixed them. ECE is the expected calibration error. Criteria 2 and 3 give the difference in log loss with its 95 percent interval from 10,000 article resamples; Gemma 4 31B-it is calibrated on validation and reads the same input as each model (4,096 tokens for the full-data model, 6,144 for the released model).}
\label{tab:bar}
\begin{tabular}{c>{\raggedright\arraybackslash}p{3.4cm}cccc}
\toprule
 & Criterion & Full-data model & Result & Released model & Result \\
\midrule
1 & ECE at most the limit & 0.0119 (limit 0.05) & met & 0.0168 (limit 0.05) & met \\[2pt]
2 & Log loss minus the class prior's, interval below 0 & \begin{tabular}[t]{@{}c@{}}$-$0.8773\\[-1pt]{\scriptsize ($-$0.9225 to $-$0.8281)}\end{tabular} & met & \begin{tabular}[t]{@{}c@{}}$-$0.8603\\[-1pt]{\scriptsize ($-$0.9078 to $-$0.8104)}\end{tabular} & met \\[2pt]
3 & Log loss minus Gemma 4 31B-it's, interval below 0 & \begin{tabular}[t]{@{}c@{}}$-$0.2084\\[-1pt]{\scriptsize ($-$0.2720 to $-$0.1505)}\end{tabular} & met & \begin{tabular}[t]{@{}c@{}}$-$0.1829\\[-1pt]{\scriptsize ($-$0.2598 to $-$0.1164)}\end{tabular} & met \\[2pt]
4 & Macro-F1 not below Gemma 4 31B-it's & 0.9399 against 0.8626 & met & 0.9248 against 0.8668 & met \\[2pt]
\bottomrule
\end{tabular}

\end{table}

\paragraph{Main results.} Table~\ref{tab:main} gives the released model beside the full-data model, Gemma 4 31B-it and the class-prior baseline, with the paired differences reported with the primary analysis. Its accuracy was 0.9269 (0.9089 to 0.9426), and its macro-F1 0.9248. Before calibration its ECE was 0.0373 (0.0281 to 0.0546) and its log loss 0.2405 (0.1826 to 0.3037) (exploratory; no interval was computed for the change).

\begin{table}[t]
\centering\footnotesize
\caption{Evidence Inference 2.0, test split (1,218 prompts in 333 articles), each model calibrated on validation. Each cell gives the estimate and its 95 percent interval from 1,000 article resamples. ECE is the expected calibration error, whose percentile interval can lie above its estimate, since the ECE is biased upward in resamples; the Brier score is summed over the three answers. Gemma 4 31B-it reads the same 6,144-token input as the released model; the full-data model's results are on its own 4,096-token input. The class prior gives every prompt the training split's label distribution. The last two rows give the paired differences reported with the primary analysis, the released model minus each reference on the same resamples; the criteria themselves are decided by Table~\ref{tab:bar}'s intervals, from 10,000 resamples. \textsuperscript{\dag}No difference was detected: the interval includes zero.}
\label{tab:main}
\setlength{\tabcolsep}{3.5pt}
\begin{tabular}{>{\raggedright\arraybackslash}p{3.0cm}ccccc}
\toprule
System & Accuracy & Macro-F1 & Log loss & Brier score & ECE \\
\midrule
Released model & \begin{tabular}[t]{@{}c@{}}0.9269\\[-1pt]{\scriptsize (0.9089 to 0.9426)}\end{tabular} & \begin{tabular}[t]{@{}c@{}}0.9248\\[-1pt]{\scriptsize (0.9059 to 0.9409)}\end{tabular} & \begin{tabular}[t]{@{}c@{}}0.2164\\[-1pt]{\scriptsize (0.1714 to 0.2645)}\end{tabular} & \begin{tabular}[t]{@{}c@{}}0.1101\\[-1pt]{\scriptsize (0.0871 to 0.1363)}\end{tabular} & \begin{tabular}[t]{@{}c@{}}0.0168\\[-1pt]{\scriptsize (0.0124 to 0.0348)}\end{tabular} \\[2pt]
Full-data model & \begin{tabular}[t]{@{}c@{}}0.9409\\[-1pt]{\scriptsize (0.9246 to 0.9554)}\end{tabular} & \begin{tabular}[t]{@{}c@{}}0.9399\\[-1pt]{\scriptsize (0.9225 to 0.9544)}\end{tabular} & \begin{tabular}[t]{@{}c@{}}0.1994\\[-1pt]{\scriptsize (0.1573 to 0.2486)}\end{tabular} & \begin{tabular}[t]{@{}c@{}}0.1001\\[-1pt]{\scriptsize (0.0778 to 0.1266)}\end{tabular} & \begin{tabular}[t]{@{}c@{}}0.0119\\[-1pt]{\scriptsize (0.0119 to 0.0300)}\end{tabular} \\[2pt]
Gemma 4 31B-it, same input & \begin{tabular}[t]{@{}c@{}}0.8752\\[-1pt]{\scriptsize (0.8489 to 0.8991)}\end{tabular} & \begin{tabular}[t]{@{}c@{}}0.8668\\[-1pt]{\scriptsize (0.8390 to 0.8935)}\end{tabular} & \begin{tabular}[t]{@{}c@{}}0.3993\\[-1pt]{\scriptsize (0.3301 to 0.4793)}\end{tabular} & \begin{tabular}[t]{@{}c@{}}0.2005\\[-1pt]{\scriptsize (0.1649 to 0.2395)}\end{tabular} & \begin{tabular}[t]{@{}c@{}}0.0283\\[-1pt]{\scriptsize (0.0238 to 0.0567)}\end{tabular} \\[2pt]
Class prior & \begin{tabular}[t]{@{}c@{}}0.4212\\[-1pt]{\scriptsize (0.3855 to 0.4584)}\end{tabular} & \begin{tabular}[t]{@{}c@{}}0.1976\\[-1pt]{\scriptsize (0.1855 to 0.2095)}\end{tabular} & \begin{tabular}[t]{@{}c@{}}1.0767\\[-1pt]{\scriptsize (1.0580 to 1.0940)}\end{tabular} & \begin{tabular}[t]{@{}c@{}}0.6525\\[-1pt]{\scriptsize (0.6395 to 0.6643)}\end{tabular} & \begin{tabular}[t]{@{}c@{}}0.0305\\[-1pt]{\scriptsize (0.0018 to 0.0663)}\end{tabular} \\[2pt]
\midrule
Released minus Gemma 4 31B-it & \begin{tabular}[t]{@{}c@{}}0.0517\\[-1pt]{\scriptsize (0.0303 to 0.0760)}\end{tabular} & \begin{tabular}[t]{@{}c@{}}0.0580\\[-1pt]{\scriptsize (0.0340 to 0.0849)}\end{tabular} & \begin{tabular}[t]{@{}c@{}}$-$0.1829\\[-1pt]{\scriptsize ($-$0.2573 to $-$0.1164)}\end{tabular} & \begin{tabular}[t]{@{}c@{}}$-$0.0904\\[-1pt]{\scriptsize ($-$0.1273 to $-$0.0570)}\end{tabular} & \begin{tabular}[t]{@{}c@{}}$-$0.0116\textsuperscript{\dag}\\[-1pt]{\scriptsize ($-$0.0345 to 0.0027)}\end{tabular} \\[2pt]
Released minus class prior & \begin{tabular}[t]{@{}c@{}}0.5057\\[-1pt]{\scriptsize (0.4656 to 0.5419)}\end{tabular} & \begin{tabular}[t]{@{}c@{}}0.7272\\[-1pt]{\scriptsize (0.7050 to 0.7479)}\end{tabular} & \begin{tabular}[t]{@{}c@{}}$-$0.8603\\[-1pt]{\scriptsize ($-$0.9102 to $-$0.8093)}\end{tabular} & \begin{tabular}[t]{@{}c@{}}$-$0.5424\\[-1pt]{\scriptsize ($-$0.5692 to $-$0.5138)}\end{tabular} & \begin{tabular}[t]{@{}c@{}}$-$0.0138\textsuperscript{\dag}\\[-1pt]{\scriptsize ($-$0.0434 to 0.0233)}\end{tabular} \\[2pt]
\bottomrule
\end{tabular}

\end{table}

\paragraph{Published results.} Table~\ref{tab:published} sets the published EI results beside the two models' macro-F1 and their intervals, as the full-data model's registration asks. Under strict scoring against each prompt's majority label, the released model's macro-F1, 0.9248, is above the best strictly scored full-text result I found, the pipeline of \citet{deyoung2020} at 0.780, and above the 0.851 of their classifier given the annotated evidence spans, which is told where the evidence lies. Lenient scoring can only raise a score, so it is also above the best leniently scored result, 0.8154 \citep{buchmann2024structure}, and it is above the 0.86 and 0.87 that \citet{buchmann2024lab} report under their own scoring. This is not a controlled comparison. The published systems read the articles in other ways, were scored and counted differently and used older base models, and the base models here may have seen EI's public test labels in pretraining (Section~\ref{sec:reading}). The gap has candidate explanations that this paper does not separate: a newer base model, the input (title and abstract first, then results passages chosen by BM25), initialization from a PubMedQA model, and recall of the test labels.

\begin{table}[t]
\centering\footnotesize
\caption{Published results on Evidence Inference 2.0's test split beside the two models' macro-F1. The published values are the papers' own; those systems read the articles in other ways and were scored and counted differently, so this is not a controlled comparison. The full-data model's figure is scored as each paper scored (strict, or lenient as \citet{buchmann2024structure} scored); the released model's is strict, since its analyses did not compute the lenient score, which can only raise it. Each interval is a 95 percent interval from 1,000 article resamples. LED is the Longformer Encoder-Decoder.}
\label{tab:published}
\setlength{\tabcolsep}{4.5pt}
\begin{tabular}{>{\raggedright\arraybackslash}p{6.0cm}lccc}
\toprule
Published system & Scoring & Its macro-F1 & \shortstack{Full-data model,\\ same scoring} & \shortstack{Released model,\\ strict} \\
\midrule
\citet{deyoung2020}, full-text pipeline & strict & 0.780 & 0.9399 {\scriptsize (0.9225 to 0.9544)} & 0.9248 {\scriptsize (0.9059 to 0.9409)} \\[1pt]
\citet{deyoung2020}, evidence spans given & strict & 0.851 & 0.9399 {\scriptsize (0.9225 to 0.9544)} & 0.9248 {\scriptsize (0.9059 to 0.9409)} \\[1pt]
\citet{buchmann2024structure}, LED, no structure infusion & lenient & 0.7430 & 0.9407 {\scriptsize (0.9240 to 0.9551)} & 0.9248 {\scriptsize (0.9059 to 0.9409)} \\[1pt]
\citet{buchmann2024structure}, LED, best setting & lenient & 0.7649 & 0.9407 {\scriptsize (0.9240 to 0.9551)} & 0.9248 {\scriptsize (0.9059 to 0.9409)} \\[1pt]
\citet{buchmann2024structure}, LongT5 & lenient & 0.8154 & 0.9407 {\scriptsize (0.9240 to 0.9551)} & 0.9248 {\scriptsize (0.9059 to 0.9409)} \\[1pt]
\bottomrule
\end{tabular}

\end{table}

\begin{figure}[t]
\begin{minipage}[c]{0.40\textwidth}
\includegraphics[width=\linewidth]{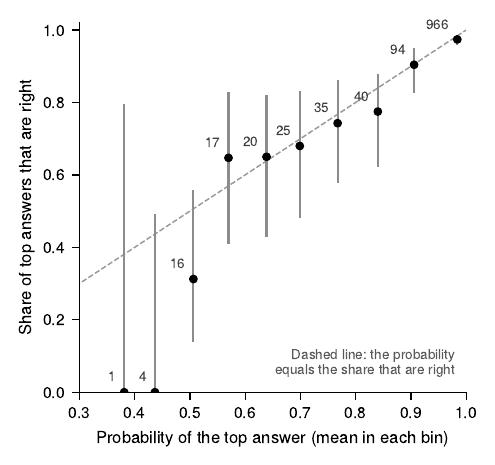}
\end{minipage}\hfill
\begin{minipage}[c]{0.56\textwidth}
\caption{Reliability diagram of the released model on Evidence Inference 2.0's test split (1,218 prompts). The prompts are grouped in the 15 equal-width bins of the reported expected calibration error by the probability of the top answer; each point is a bin's mean probability against the share of its top answers that were right, with the bin's count of prompts and a 95 percent Wilson interval (exploratory). The Wilson intervals treat the prompts as independent, so they are narrower than the article-level intervals used elsewhere. Bins with no prompt are not drawn.}
\label{fig:reliability}
\end{minipage}
\end{figure}

\paragraph{Calibration.} Figure~\ref{fig:reliability} shows the released model's reliability diagram on the test split, in the 15 bins of the reported ECE. Most prompts, 966, fall in the top bin (probabilities above 0.9333), where the share of top answers that were right was 0.9741 against a mean probability of 0.9840 (exploratory). For each class, no miscalibration was detected by its calibration slope and intercept (Appendix~\ref{app:answers}).

\paragraph{Other secondary analyses.} On the permissively licensed validation prompts, the longer input's two-run average scored a lower log loss than the baseline configuration's, by 0.0286 (0.0097 to 0.0496); for the other candidates no difference was detected (Appendix~\ref{app:runs}). On test, the released model minus the baseline configuration's three-run ensemble gave a log-loss difference of 0.0000 ($-$0.0202 to 0.0174): no difference was detected. The soup's log loss was higher than that of the three runs' prediction ensemble by 0.0048 (0.0010 to 0.0092). The four-method calibration rule chose the same method as the full-data model's two-method rule, so the difference between the two rules is zero. Swap test-time augmentation raised log loss by 0.0922 (0.0672 to 0.1206) and ECE by 0.0481 (0.0259 to 0.0602). Appendix~\ref{app:other} reports the PubMedQA and readmission models.

\paragraph{Reproducibility check.} I ran the release as a user would, at the fixed revision that the statements give: its own inference script on a MacBook Pro with an Apple M5 chip, under macOS, with each task's examples in a file of their own; the language models ran on the laptop's graphics processing unit (GPU) and the readmission trees on its central processing unit (CPU). Rules fixed before the check's first run compared the outputs with the evaluated test predictions: readmission passes when every probability is within 0.0001 of the evaluated one, and EI and PubMedQA pass when no top answer changes on an example whose evaluated top probability is at least 0.6 and the mean absolute difference over all examples and answers is at most 0.02. All three passed. On the 6,998 readmission encounters, the largest difference was below one in ten million. On the 500 PubMedQA items, the mean absolute difference was 0.0013 and the largest 0.0159; on 60 EI prompts drawn at random from the 1,218, each with the input it was evaluated on, the mean was 0.0028 and the largest 0.0459. No top answer changed on either. The EI examples hold each article as the evaluated predictions read it, already cut to 6,144 tokens, so the check does not cover the passage selection of a whole article.

\section{Does the model use the article?}
\label{sec:reading}

\begin{figure}[t]
\begin{minipage}[c]{0.46\textwidth}
\includegraphics[width=\linewidth]{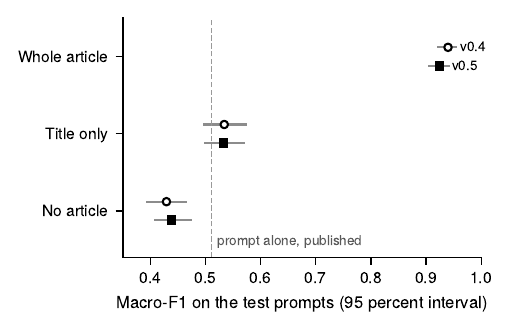}
\end{minipage}\hfill
\begin{minipage}[c]{0.50\textwidth}
\caption{Macro-F1 of the full-data model (v0.4, open circles) and the released model (v0.5, filled squares) on Evidence Inference 2.0's test prompts, with the whole article (1,218 prompts), the title only (1,209) and no article (1,218), each with its 95 percent interval from 1,000 article resamples. The dashed line is the macro-F1 that \citet{deyoung2020} report for a classifier given the prompt alone, 0.511 (a different model, scored and counted as that paper did).}
\label{fig:reading}
\end{minipage}
\end{figure}

EI's labels have been public since 2019, the trials' results are in PubMed, and the pretraining text of Qwen3 and Gemma 4 is not published. A model could answer some prompts from what it saw in pretraining rather than from the article in front of it. Two context ablations ask the test prompts with no article and with the title alone, and a counterfactual test swaps the intervention and the comparator; the interpretations of all three were fixed in the amendment before they ran (Table~\ref{tab:reading}, Figure~\ref{fig:reading}).

\begin{table}[t]
\centering\footnotesize
\caption{Context ablations and the swap test on Evidence Inference 2.0's test split (pre-registered, with interpretations fixed in advance). Each cell gives the estimate and its 95 percent interval from 1,000 article resamples. The title-only input leaves out the 9 prompts whose article gives no title; the title's difference is computed on the prompts in both inputs. ``Direction answers'' are ``significantly increased'' and ``significantly decreased''; for them the swapped prompt's expected answer is the other direction.}
\label{tab:reading}
\begin{tabular}{>{\raggedright\arraybackslash}p{5.4cm}rcrc}
\toprule
Statistic & Prompts & Full-data model & Prompts & Released model \\
\midrule
\multicolumn{5}{@{}l}{\textit{No article}} \\[1pt]
Accuracy & 1,218 & 0.5090 {\scriptsize (0.4779 to 0.5439)} & 1,218 & 0.5156 {\scriptsize (0.4874 to 0.5494)} \\
Macro-F1 & 1,218 & 0.4289 {\scriptsize (0.3941 to 0.4634)} & 1,218 & 0.4384 {\scriptsize (0.4085 to 0.4726)} \\
Agreement with the whole-article answer & 1,218 & 0.5140 {\scriptsize (0.4812 to 0.5473)} & 1,218 & 0.5238 {\scriptsize (0.4955 to 0.5571)} \\
\multicolumn{5}{@{}l}{\textit{Title only}} \\[1pt]
Accuracy & 1,209 & 0.5715 {\scriptsize (0.5346 to 0.6065)} & 1,209 & 0.5715 {\scriptsize (0.5389 to 0.6041)} \\
Macro-F1 & 1,209 & 0.5338 {\scriptsize (0.4964 to 0.5733)} & 1,209 & 0.5327 {\scriptsize (0.4995 to 0.5692)} \\
Agreement with the whole-article answer & 1,209 & 0.5682 {\scriptsize (0.5343 to 0.6027)} & 1,209 & 0.5782 {\scriptsize (0.5471 to 0.6108)} \\
Macro-F1, no article, on the same prompts & 1,209 & 0.4314 {\scriptsize (0.3967 to 0.4650)} & 1,209 & 0.4388 {\scriptsize (0.4076 to 0.4723)} \\
Macro-F1, title only minus no article & 1,209 & 0.1024 {\scriptsize (0.0736 to 0.1331)} & 1,209 & 0.0939 {\scriptsize (0.0655 to 0.1234)} \\
\multicolumn{5}{@{}l}{\textit{Intervention and comparator swapped}} \\[1pt]
Accuracy against the swapped labels & 1,218 & 0.8103 {\scriptsize (0.7802 to 0.8400)} & 1,218 & 0.7808 {\scriptsize (0.7493 to 0.8117)} \\
Macro-F1 against the swapped labels & 1,218 & 0.7958 {\scriptsize (0.7633 to 0.8283)} & 1,218 & 0.7600 {\scriptsize (0.7268 to 0.7946)} \\
Share of direction answers that flip & 703 & 0.7226 {\scriptsize (0.6765 to 0.7714)} & 699 & 0.6652 {\scriptsize (0.6208 to 0.7157)} \\
Share of no-difference answers that stay & 515 & 0.9806 {\scriptsize (0.9657 to 0.9931)} & 519 & 0.9672 {\scriptsize (0.9477 to 0.9841)} \\
\multicolumn{5}{@{}l}{\textit{Swapped, on the prompts whose comparator is a time point or a baseline}} \\[1pt]
Share of direction answers that flip & 38 & 0.1316 {\scriptsize (0.0250 to 0.2632)} & 37 & 0.2162 {\scriptsize (0.0857 to 0.3439)} \\
Share of no-difference answers that stay & 10 & 1.0000 {\scriptsize (1.0000 to 1.0000)} & 11 & 1.0000 {\scriptsize (1.0000 to 1.0000)} \\
\bottomrule
\end{tabular}

\end{table}

\paragraph{No article.} With the article replaced by empty text, the released model's macro-F1 was 0.4384 (0.4085 to 0.4726), and the full-data model's 0.4289 (0.3941 to 0.4634), against 0.1976 (0.1855 to 0.2095) for the class-prior baseline, a reference the amendment fixed. The pre-specified interpretation of a macro-F1 at or below 0.60 is that the answers depend on the article's text, and it applies to both. This shows that the models need the article, not that they never saw the test set: a model always trained with articles can fail without one either way. Both intervals lie below the 0.511 that \citet{deyoung2020} report for a classifier trained to answer from the prompt alone, a different model on their own scoring.

\paragraph{Title only.} On the 1,209 prompts whose article gives a title, the title alone raised the released model's macro-F1 over no article, on the same prompts, by 0.0939 (0.0655 to 0.1234), and the full-data model's by 0.1024 (0.0736 to 0.1331). The intervals lie wholly above 0.05, so the pre-specified interpretation applies to both: ``the title adds information about the result: a title can state the result, or the model may recall the trial, and these checks cannot tell them apart''. \citet{windisch2026} report that frontier models reached an accuracy of 0.79 to 0.88 from titles alone on 250 oncology trials, on a two-answer task whose majority class was 0.584.

\paragraph{Intervention and comparator swapped.} When the intervention and the comparator are swapped, a model that follows the comparison as written should flip its ``increased'' and ``decreased'' answers. The released model flipped 0.6652 (0.6208 to 0.7157) of its direction answers, and the full-data model 0.7226 (0.6765 to 0.7714). Where the comparator is a time point or a baseline, the released model flipped 0.2162 of its 37 direction answers. Against the swapped labels, the released model's accuracy was 0.7808 (0.7493 to 0.8117) and the full-data model's 0.8103 (0.7802 to 0.8400), which the pre-specified interpretation calls ``partly'' for both. This is a weakness: the model often keeps its direction when the comparison is reversed. A fix to test is to train on prompts whose comparator is a time point or a baseline together with their swapped copies, or to add a consistency loss that asks the answer to a swapped prompt to be the flip of the original answer. Swap augmentation of all training prompts was a candidate in model selection and did not pass, but selection measured validation log loss, not consistency under the swap; swap test-time augmentation made log loss worse (Section~\ref{sec:results}).

\section{The cost of license filtering}
\label{sec:license}

This pre-registered secondary analysis compares fine-tuning on the permissively licensed articles alone with fine-tuning on all of them. The baseline configuration, fine-tuned on the 1,431 permissively licensed training articles and calibrated on the 654 permissively licensed validation prompts (three runs, their probabilities averaged), had a lower accuracy than the full-data model, fine-tuned on all 2,657 and calibrated on all 1,237 validation prompts, by 0.0123 (0.0034 to 0.0207), and a higher log loss by 0.0170 (0.0058 to 0.0284) (Table~\ref{tab:k2}). Both chose temperature scaling, which does not change the top answer, so the accuracy difference comes from training alone, while the log-loss difference also reflects the different calibration prompts. The released model, which also uses the longer input, had a lower accuracy than the full-data model by 0.0140 (0.0025 to 0.0245); for log loss no difference was detected, 0.0169 ($-$0.0095 to 0.0397).

\begin{table}[t]
\centering\footnotesize
\caption{The cost of license filtering: each model minus the full-data model on Evidence Inference 2.0's test split (1,218 prompts in 333 articles), with 95 percent intervals from 1,000 article resamples. Each is calibrated on validation: the models fine-tuned on permissively licensed articles on the 654 permissively licensed validation prompts, the full-data model on all 1,237. ECE is the expected calibration error. \textsuperscript{\dag}No difference was detected: the interval includes zero.}
\label{tab:k2}
\setlength{\tabcolsep}{3.5pt}
\begin{tabular}{>{\raggedright\arraybackslash}p{3.0cm}ccccc}
\toprule
Minus the full-data model & Accuracy & Macro-F1 & Log loss & Brier score & ECE \\
\midrule
Same configuration, permissively licensed articles only & \begin{tabular}[t]{@{}c@{}}$-$0.0123\\[-1pt]{\scriptsize ($-$0.0207 to $-$0.0034)}\end{tabular} & \begin{tabular}[t]{@{}c@{}}$-$0.0133\\[-1pt]{\scriptsize ($-$0.0226 to $-$0.0041)}\end{tabular} & \begin{tabular}[t]{@{}c@{}}0.0170\\[-1pt]{\scriptsize (0.0058 to 0.0284)}\end{tabular} & \begin{tabular}[t]{@{}c@{}}0.0106\\[-1pt]{\scriptsize (0.0037 to 0.0174)}\end{tabular} & \begin{tabular}[t]{@{}c@{}}0.0007\textsuperscript{\dag}\\[-1pt]{\scriptsize ($-$0.0105 to 0.0102)}\end{tabular} \\[2pt]
Released model & \begin{tabular}[t]{@{}c@{}}$-$0.0140\\[-1pt]{\scriptsize ($-$0.0245 to $-$0.0025)}\end{tabular} & \begin{tabular}[t]{@{}c@{}}$-$0.0152\\[-1pt]{\scriptsize ($-$0.0261 to $-$0.0036)}\end{tabular} & \begin{tabular}[t]{@{}c@{}}0.0169\textsuperscript{\dag}\\[-1pt]{\scriptsize ($-$0.0095 to 0.0397)}\end{tabular} & \begin{tabular}[t]{@{}c@{}}0.0100\textsuperscript{\dag}\\[-1pt]{\scriptsize ($-$0.0055 to 0.0231)}\end{tabular} & \begin{tabular}[t]{@{}c@{}}0.0048\textsuperscript{\dag}\\[-1pt]{\scriptsize ($-$0.0093 to 0.0144)}\end{tabular} \\[2pt]
\bottomrule
\end{tabular}

\end{table}

\section{Selective prediction and deferral}
\label{sec:handoff}

This pre-registered secondary analysis has two parts: the released model's risk-coverage curve, which shows how many of its answers a reviewer could accept without checking them, and deferral of its least confident answers to Gemma 4 31B-it given the same input; the released model itself defers nothing. On the curve, accepting the 80 percent most confident answers (975 of the 1,218 test prompts, those kept at the pre-registered deferral rate of 20 percent) leaves 25 errors among them, an error rate of 0.0256, against 0.0731 for all answers, and the 35 percent most confident (427 prompts) hold no error; these points are read from the pre-registered curve, and no interval was computed for them. The area under the released model's risk-coverage curve (AURC) was 0.0128 (0.0082 to 0.0182), its excess over a perfect ranking of the same answers 0.0100 (0.0064 to 0.0145), and the area under the generalized risk-coverage curve 0.0105 (0.0069 to 0.0147); the curves at every 5 percent of coverage are in the results files.

Table~\ref{tab:handoff} gives the combined accuracy at the four pre-registered deferral rates for five ways of choosing the prompts to defer. Deferring the answers with the lowest top probability gave a combined accuracy of 0.9335 at 10 percent deferred and 0.8768 at 50 percent, against 0.9269 for the released model alone and 0.8752 for Gemma 4 31B-it alone; the oracle's bound was 0.9573. The results files hold no interval for the cascade minus the released model alone, so no difference from the released model alone is claimed. The released model was wrong where Gemma 4 31B-it was right on 37 of the 1,218 test prompts (a share of 0.0304), and right where Gemma 4 31B-it was wrong on 100: these few prompts bound what any deferral rule could gain and leave a learned rule few cases to learn from. Against deferral by top probability, no difference was detected for the rule learned on validation at any rate, and every interval of that difference lies between a loss of 0.0048 and a gain of 0.0041 in combined accuracy. For the full-data model with Gemma 4 31B-it on 4,096-token inputs, the combined accuracy at 10 percent was 0.9294 against 0.9409 alone; for the PubMedQA model, 0.8000 against 0.7960 alone.

\begin{table}[t]
\centering\scriptsize
\caption{Deferral (pre-registered, secondary): the combined accuracy of the released model and Gemma 4 31B-it, given the same input, when a share of the 1,218 test prompts is deferred to Gemma 4 31B-it, with 95 percent intervals from 1,000 article resamples. Max probability defers first the answers whose top probability is lowest. The learned rule was fitted and frozen on validation. Random gives the expected value of random deferral. Gemma 4 31B-it's confidence defers first the items on which its own top probability is highest, which needs its answer on every item. The oracle defers the items the released model gets wrong and Gemma 4 31B-it gets right, which bounds any rule. The released model alone: 0.9269; Gemma 4 31B-it alone: 0.8752.}
\label{tab:handoff}
\begin{tabular}{lrccccc}
\toprule
Deferred & Prompts & Max probability & Learned rule & Random (expected) & \shortstack{Gemma 4 31B-it's\\ confidence} & Oracle \\
\midrule
10 percent & 121 & \begin{tabular}[t]{@{}c@{}}0.9335\\[-1pt]{\scriptsize (0.9160 to 0.9492)}\end{tabular} & \begin{tabular}[t]{@{}c@{}}0.9335\\[-1pt]{\scriptsize (0.9150 to 0.9493)}\end{tabular} & \begin{tabular}[t]{@{}c@{}}0.9218\\[-1pt]{\scriptsize (0.9040 to 0.9369)}\end{tabular} & \begin{tabular}[t]{@{}c@{}}0.9269\\[-1pt]{\scriptsize (0.9097 to 0.9428)}\end{tabular} & \begin{tabular}[t]{@{}c@{}}0.9573\\[-1pt]{\scriptsize (0.9442 to 0.9693)}\end{tabular} \\[2pt]
20 percent & 243 & \begin{tabular}[t]{@{}c@{}}0.9171\\[-1pt]{\scriptsize (0.8986 to 0.9328)}\end{tabular} & \begin{tabular}[t]{@{}c@{}}0.9163\\[-1pt]{\scriptsize (0.8993 to 0.9336)}\end{tabular} & \begin{tabular}[t]{@{}c@{}}0.9166\\[-1pt]{\scriptsize (0.8988 to 0.9322)}\end{tabular} & \begin{tabular}[t]{@{}c@{}}0.9261\\[-1pt]{\scriptsize (0.9076 to 0.9430)}\end{tabular} & \begin{tabular}[t]{@{}c@{}}0.9573\\[-1pt]{\scriptsize (0.9442 to 0.9693)}\end{tabular} \\[2pt]
30 percent & 365 & \begin{tabular}[t]{@{}c@{}}0.8966\\[-1pt]{\scriptsize (0.8751 to 0.9173)}\end{tabular} & \begin{tabular}[t]{@{}c@{}}0.8966\\[-1pt]{\scriptsize (0.8750 to 0.9170)}\end{tabular} & \begin{tabular}[t]{@{}c@{}}0.9114\\[-1pt]{\scriptsize (0.8935 to 0.9276)}\end{tabular} & \begin{tabular}[t]{@{}c@{}}0.9245\\[-1pt]{\scriptsize (0.9055 to 0.9423)}\end{tabular} & \begin{tabular}[t]{@{}c@{}}0.9573\\[-1pt]{\scriptsize (0.9442 to 0.9693)}\end{tabular} \\[2pt]
50 percent & 609 & \begin{tabular}[t]{@{}c@{}}0.8768\\[-1pt]{\scriptsize (0.8535 to 0.9013)}\end{tabular} & \begin{tabular}[t]{@{}c@{}}0.8777\\[-1pt]{\scriptsize (0.8532 to 0.9014)}\end{tabular} & \begin{tabular}[t]{@{}c@{}}0.9011\\[-1pt]{\scriptsize (0.8821 to 0.9182)}\end{tabular} & \begin{tabular}[t]{@{}c@{}}0.9261\\[-1pt]{\scriptsize (0.9073 to 0.9441)}\end{tabular} & \begin{tabular}[t]{@{}c@{}}0.9573\\[-1pt]{\scriptsize (0.9442 to 0.9693)}\end{tabular} \\[2pt]
\bottomrule
\end{tabular}

\end{table}

The released model's and Gemma 4 31B-it's EI test predictions ran on a rented NVIDIA H100 SXM GPU with 80 GB, at \$3.52 an hour; at that rate, 1,000 EI questions cost \$0.297 of GPU time with the released model and \$2.443 with Gemma 4 31B-it, from the time each took on the GPU (Appendix~\ref{app:cost}).

\section{Limitations}
\label{sec:limits}

\begin{itemize}\setlength{\itemsep}{1pt}
\item For trials, one task was tested: one comparison in a randomized trial's article, in English, with the article cut to 6,144 tokens. Other questions, wordings and kinds of document are untested, and calibration under such a shift in distribution is unmeasured.
\item EI's test split was used for the full-data model's criteria, secondary analyses and context ablations before the released model's criteria; PubMedQA's and readmission's test splits were used in every version.
\item The pretraining text of Qwen3 and Gemma 4 is not published and may include PMC articles (EI's articles come from PMC), EI's repository (public since 2019), PubMedQA and the readmission data; this would favor both models. EI's test labels are public, so no evaluation of EI can be blind. The PubMedQA model's pseudo-labels came from a model further pretrained on PubMed abstracts (Appendix~\ref{app:other}), which may include the abstracts of EI's test articles but no EI label. In an exploratory check, 10 EI test articles (37 prompts) shared sentences with PubMedQA abstracts that the PubMedQA model, from which the EI models were initialized, had seen in training; the full-data model's accuracy on them was 0.8378 (0.6667 to 0.9445), and no difference from the other prompts was detected, $-$0.1063 ($-$0.2817 to 0.0051).
\item The title effect is unresolved, and the model often keeps its direction when the intervention and comparator are swapped (Section~\ref{sec:reading}).
\item The released model's EI fine-tuning used only permissively licensed articles, with the two caveats of Section~\ref{sec:data}: the passage selection's word statistics, counted over all training articles, and its initialization from the PubMedQA model, whose abstracts' licenses were not checked one by one.
\item The baselines are limited. Gemma 4 31B-it is a reference that was not trained on the task: it ran zero-shot with thinking off, its answers read from its next-token letter probabilities, and its prompt template and that reading were fixed, not tuned. No fine-tuned encoder and no few-shot or reasoning-enabled large model was run on the same inputs, and no ablation isolates the input format or the PubMedQA initialization.
\item The model returns no evidence span, which would let a reviewer check an answer against the article.
\item Fiorillo is a research model and is not for clinical use. No person was recruited for this work.
\end{itemize}

\section{Discussion}
\label{sec:discussion}

A 4-billion-parameter open model, fine-tuned for EI only on articles under CC BY, CC0 or the public domain (it was initialized from a PubMedQA model whose abstracts' licenses were not checked one by one), met criteria fixed in advance on EI's test split, in the split's second pre-registered use. Its stated probabilities matched how often it was right within the pre-registered limit: its expected calibration error was 0.0168 against 0.05, and no miscalibration of any answer was detected by the per-answer calibration slopes and intercepts (Appendix~\ref{app:answers}). A reviewer who accepted its 80 percent most confident answers would have met an error rate of 0.0256 among them. These figures hold as far as EI's test split represents a reviewer's trials.

What the evaluation does not show is how much of the accuracy comes from reading the article and how much from what the base model saw in pretraining. The answers depend on the article, but the title adds information, and the ablations here cannot separate a title that states the result from recall of the trial. Four results would change that reading: a test on trials published after the base model's release, where recall is not possible; membership-inference tests on the base model; counterfactual edits of EI's annotated evidence spans, which ask whether the answer follows the edited text; and a model that follows a swapped comparison.

License filtering had a measurable cost: the same configuration was less accurate when it learned only from permissively licensed articles (Table~\ref{tab:k2}). The release makes that trade explicit, and its NOTICE file credits each of the 1,610 permissively licensed training and validation articles with its identifier, license and citation.

\section*{Statements}
\label{sec:statements}

\paragraph{Conflicts of interest.} I may later build a company on this model.

\paragraph{Funding.} None. I paid for the computing myself.

\paragraph{Ethics.} This work used public, de-identified data and published articles only. No person was recruited or contacted.

\paragraph{Data and code availability.} The release: Fiorillo v0.5 on Hugging Face, digital object identifier (DOI) \doi{10.57967/hf/10722} \citep{fiorillo05}, under the Apache License 2.0, with its model card, its inference script and a NOTICE file that credits each permissively licensed article with its identifier, license and citation. Its revision \idn{0058ed7} differs from the revision the DOI records, \idn{56e5944}, only in how the script loads its libraries, so that each task runs on macOS when its examples are in a file of their own, and in a note in the model card. The earlier encoder, Fiorillo reader v1: DOI \doi{10.57967/hf/10658} \citep{reader1}. The registration and its amendment: \mbox{\url{https://osf.io/kaxmn/}} \citep{osfreg}. The code, the results files and this paper's sources: \mbox{\url{https://github.com/johann-e-li/fiorillo}}, under the Apache License 2.0.

\paragraph{Use of artificial intelligence tools.} Claude, by Anthropic, wrote much of the code, ran the analyses with me and drafted the text. I reviewed every part and take responsibility for all of it. Artificial intelligence (AI) tools are not authors.

\clearpage
\appendix

\section{Every run and model}
\label{app:runs}

Table~\ref{tab:runs} lists every run of the released model's training configurations and every model built from them, with the full-data model for reference, under the identifiers of the results files. The validation score is the model-selection score: the out-of-fold log loss of the calibrated probabilities on the 654 permissively licensed validation prompts. The margin was 0.0110. The longer input's two runs scored 0.1834 on average against the baseline configuration's 0.2116 for its first two runs, and both of its runs scored below the baseline run of the same seed; it was the only candidate that passed. The pre-registered comparison of each candidate's two-run average with the baseline's on the same prompts, from the calibrated out-of-fold probabilities, gave these log-loss differences: for the longer input, $-$0.0286 ($-$0.0496 to $-$0.0097); for swap augmentation, 0.0014 ($-$0.0156 to 0.0187); for the first higher learning rate, $-$0.0060 ($-$0.0181 to 0.0057); and for the second, $-$0.0156 ($-$0.0316 to 0.0021). For the last three no difference was detected. The longer input's third run, \sysname{s9\_long\_3}, was not calibrated when it was trained; it was calibrated afterward by the same method on the same 654 prompts, and its figures are exploratory: its validation score was 0.1823, its test log loss 0.2139 (0.1725 to 0.2626) and its accuracy 0.9319 (0.9154 to 0.9467). Its own calibration enters no frozen model or pre-registered comparison; its weights are part of the released soup, \sysname{s9\_ei\_wavg}, as built.

\begin{table}[H]
\centering\scriptsize
\caption{Every run and model of the released model's Evidence Inference 2.0 training configurations, under the identifiers of the results files (the released model is \texttt{s9\_ei\_wavg}). The validation score is the out-of-fold log loss of the calibrated probabilities on the 654 permissively licensed validation prompts (lower is better); the test columns give the estimate and its 95 percent interval from 1,000 article resamples on the 1,218 test prompts, after each model's own calibration. ECE is the expected calibration error, whose percentile interval can lie above its estimate, since the ECE is biased upward in resamples. \texttt{s9\_b} and \texttt{s9\_ei} average their runs' probabilities (prediction ensembles); \texttt{s9\_ei\_wavg} averages their weights (a uniform soup); \texttt{s9\_ei\_wavg\_sa} adds swap test-time augmentation. Runs that were not evaluated on the test split are marked so. \textsuperscript{a}The full-data model's score, on all 1,237 validation prompts, is not on the same prompts; that it matches the model's test log loss to four decimals is a coincidence (0.199444 and 0.199449). \textsuperscript{b}Calibrated afterward by the same method (exploratory); model selection did not score this run alone.}
\label{tab:runs}
\begin{tabular}{llcccc}
\toprule
Configuration & Results-file ID & Validation score & Test log loss & Test accuracy & Test ECE \\
\midrule
Baseline & \texttt{s9\_b\_1} & 0.2171 & \begin{tabular}[t]{@{}c@{}}0.2239\\[-1pt]{\scriptsize (0.1789 to 0.2746)}\end{tabular} & \begin{tabular}[t]{@{}c@{}}0.9269\\[-1pt]{\scriptsize (0.9093 to 0.9438)}\end{tabular} & \begin{tabular}[t]{@{}c@{}}0.0153\\[-1pt]{\scriptsize (0.0107 to 0.0321)}\end{tabular} \\[2pt]
 & \texttt{s9\_b\_2} & 0.2061 & \begin{tabular}[t]{@{}c@{}}0.2248\\[-1pt]{\scriptsize (0.1793 to 0.2731)}\end{tabular} & \begin{tabular}[t]{@{}c@{}}0.9245\\[-1pt]{\scriptsize (0.9058 to 0.9412)}\end{tabular} & \begin{tabular}[t]{@{}c@{}}0.0118\\[-1pt]{\scriptsize (0.0101 to 0.0306)}\end{tabular} \\[2pt]
 & \texttt{s9\_b\_3} & 0.2094 & \begin{tabular}[t]{@{}c@{}}0.2189\\[-1pt]{\scriptsize (0.1735 to 0.2704)}\end{tabular} & \begin{tabular}[t]{@{}c@{}}0.9278\\[-1pt]{\scriptsize (0.9099 to 0.9436)}\end{tabular} & \begin{tabular}[t]{@{}c@{}}0.0060\\[-1pt]{\scriptsize (0.0093 to 0.0282)}\end{tabular} \\[2pt]
 & \texttt{s9\_b} & 0.2067 & \begin{tabular}[t]{@{}c@{}}0.2164\\[-1pt]{\scriptsize (0.1714 to 0.2663)}\end{tabular} & \begin{tabular}[t]{@{}c@{}}0.9286\\[-1pt]{\scriptsize (0.9116 to 0.9446)}\end{tabular} & \begin{tabular}[t]{@{}c@{}}0.0126\\[-1pt]{\scriptsize (0.0105 to 0.0328)}\end{tabular} \\[2pt]
\midrule
Swap augmentation & \texttt{s9\_swap\_1} & 0.2100 & not evaluated & not evaluated & not evaluated \\[2pt]
 & \texttt{s9\_swap\_2} & 0.2174 & not evaluated & not evaluated & not evaluated \\[2pt]
\midrule
First higher learning rate & \texttt{s9\_lr1\_1} & 0.2153 & not evaluated & not evaluated & not evaluated \\[2pt]
 & \texttt{s9\_lr1\_2} & 0.2015 & not evaluated & not evaluated & not evaluated \\[2pt]
\midrule
Second higher learning rate & \texttt{s9\_lr2\_1} & 0.2053 & not evaluated & not evaluated & not evaluated \\[2pt]
 & \texttt{s9\_lr2\_2} & 0.1983 & not evaluated & not evaluated & not evaluated \\[2pt]
\midrule
Longer inputs (chosen) & \texttt{s9\_long\_1} & 0.1816 & \begin{tabular}[t]{@{}c@{}}0.2177\\[-1pt]{\scriptsize (0.1728 to 0.2671)}\end{tabular} & \begin{tabular}[t]{@{}c@{}}0.9245\\[-1pt]{\scriptsize (0.9072 to 0.9408)}\end{tabular} & \begin{tabular}[t]{@{}c@{}}0.0215\\[-1pt]{\scriptsize (0.0137 to 0.0374)}\end{tabular} \\[2pt]
 & \texttt{s9\_long\_2} & 0.1852 & \begin{tabular}[t]{@{}c@{}}0.2235\\[-1pt]{\scriptsize (0.1794 to 0.2738)}\end{tabular} & \begin{tabular}[t]{@{}c@{}}0.9294\\[-1pt]{\scriptsize (0.9118 to 0.9447)}\end{tabular} & \begin{tabular}[t]{@{}c@{}}0.0226\\[-1pt]{\scriptsize (0.0139 to 0.0392)}\end{tabular} \\[2pt]
 & \texttt{s9\_long\_3}\textsuperscript{b} & 0.1823 & \begin{tabular}[t]{@{}c@{}}0.2139\\[-1pt]{\scriptsize (0.1725 to 0.2626)}\end{tabular} & \begin{tabular}[t]{@{}c@{}}0.9319\\[-1pt]{\scriptsize (0.9154 to 0.9467)}\end{tabular} & \begin{tabular}[t]{@{}c@{}}0.0164\\[-1pt]{\scriptsize (0.0118 to 0.0328)}\end{tabular} \\[2pt]
 & \texttt{s9\_ei} & 0.1790 & \begin{tabular}[t]{@{}c@{}}0.2116\\[-1pt]{\scriptsize (0.1692 to 0.2594)}\end{tabular} & \begin{tabular}[t]{@{}c@{}}0.9302\\[-1pt]{\scriptsize (0.9125 to 0.9457)}\end{tabular} & \begin{tabular}[t]{@{}c@{}}0.0115\\[-1pt]{\scriptsize (0.0103 to 0.0316)}\end{tabular} \\[2pt]
 & \texttt{s9\_ei\_wavg} & 0.1781 & \begin{tabular}[t]{@{}c@{}}0.2164\\[-1pt]{\scriptsize (0.1714 to 0.2645)}\end{tabular} & \begin{tabular}[t]{@{}c@{}}0.9269\\[-1pt]{\scriptsize (0.9089 to 0.9426)}\end{tabular} & \begin{tabular}[t]{@{}c@{}}0.0168\\[-1pt]{\scriptsize (0.0124 to 0.0348)}\end{tabular} \\[2pt]
 & \texttt{s9\_ei\_wavg\_sa} & 0.2515 & \begin{tabular}[t]{@{}c@{}}0.3086\\[-1pt]{\scriptsize (0.2545 to 0.3692)}\end{tabular} & \begin{tabular}[t]{@{}c@{}}0.9212\\[-1pt]{\scriptsize (0.9037 to 0.9380)}\end{tabular} & \begin{tabular}[t]{@{}c@{}}0.0648\\[-1pt]{\scriptsize (0.0510 to 0.0818)}\end{tabular} \\[2pt]
\midrule
Full-data model & \texttt{s8\_ei} & 0.1994\textsuperscript{a} & \begin{tabular}[t]{@{}c@{}}0.1994\\[-1pt]{\scriptsize (0.1573 to 0.2486)}\end{tabular} & \begin{tabular}[t]{@{}c@{}}0.9409\\[-1pt]{\scriptsize (0.9246 to 0.9554)}\end{tabular} & \begin{tabular}[t]{@{}c@{}}0.0119\\[-1pt]{\scriptsize (0.0119 to 0.0300)}\end{tabular} \\[2pt]
\bottomrule
\end{tabular}

\end{table}

\section{The pre-registered analyses and the deviations}
\label{app:deviations}

Table~\ref{tab:analyses} maps each analysis to its code in the registration (H1 to H4 for the full-data model; K1 to K10 in the amendment), its status in this paper and the label the registration gave it.

\begin{table}[H]
\centering\footnotesize
\caption{Each analysis, its code in the registration, its status in this paper, the label the registration gave it, and where this paper reports it. The registration calls an analysis descriptive when it reports it without a test; ``none'' means it gave no label.}
\label{tab:analyses}
\setlength{\tabcolsep}{4pt}
\begin{tabular}{>{\raggedright\arraybackslash}p{5.5cm}ll>{\raggedright\arraybackslash}p{2.5cm}>{\raggedright\arraybackslash}p{2.7cm}}
\toprule
Analysis & Code & Status here & Registration's label & Reported in \\
\midrule
The four criteria, full-data model & H1 to H4 & primary & hypotheses & Table~\ref{tab:bar} \\[1pt]
The four criteria, released model, with its paired differences from Gemma 4 31B-it and the class-prior baseline, overall and for each class & K1 & primary & none & Tables~\ref{tab:bar}, \ref{tab:main} and~\ref{tab:k1answers} \\[1pt]
The cost of license filtering & K2 & secondary & descriptive & Section~\ref{sec:license} \\[1pt]
Candidate configurations against the baseline on validation; the released model against the baseline configuration on test & K3 & secondary & none & Section~\ref{sec:results}, Appendix~\ref{app:runs} \\[1pt]
Context ablations and the swap test & K4 & secondary & none; interpretations fixed in advance & Section~\ref{sec:reading} \\[1pt]
PubMedQA comparisons & K5 & secondary & descriptive & Appendix~\ref{app:other} \\[1pt]
Readmission ensembles and refitted trees & K6 & secondary & descriptive & Appendix~\ref{app:other} \\[1pt]
Deferral and risk-coverage & K7 & secondary & secondary, descriptive & Section~\ref{sec:handoff} \\[1pt]
Uniform soup against prediction ensemble & K8 & secondary & descriptive & Section~\ref{sec:results} \\[1pt]
Four calibration methods against two & K9 & secondary & descriptive & Sections~\ref{sec:model} and~\ref{sec:results} \\[1pt]
Swap test-time augmentation & K10 & secondary & descriptive & Section~\ref{sec:results} \\[1pt]
Sensitivity analysis without the flagged test prompts & no code & secondary & descriptive & Section~\ref{sec:results} \\[1pt]
Zero-probability check & no code & secondary & descriptive & Section~\ref{sec:results} \\[1pt]
Per-class calibration slopes and intercepts & no code & secondary & listed among the results to report & Appendix~\ref{app:answers} \\[1pt]
Comparison with published results, full-data model & no code & secondary & secondary, descriptive & Table~\ref{tab:published} \\[1pt]
Analyses marked exploratory & none & exploratory & not registered & where they appear \\
\bottomrule
\end{tabular}
\end{table}

Every planned change was recorded with its reason before the work it changed, except one that recorded a GPU already deployed. The changes planned before the registration was filed are listed in it, and none changed the four criteria, the calibration rule or the list of comparisons. Six of them concerned how Gemma 4 31B-it's serving engine (vLLM) was installed, checked and run. One set how Gemma 4 31B-it's answer probabilities are obtained when an option letter is missing from the top 20 log-probabilities of its first output (its top 1,000 are read first, then a second process is run; the quantity read stays its next-token probabilities); it is an input to criteria 3 and 4. The changes planned before the amendment was filed are part of what it registered; among them are two runs for each candidate configuration, deciding swap test-time augmentation on validation, no larger base model, fallback settings for BioGPT-Large-PubMedQA in case its checkpoint arguments could not be read, the title-only input's rule, three rules of the license filtering (fixed with no article's class changed), and the cost per 1,000 questions. Table~\ref{tab:deviations} lists the deviations after the registration and after the amendment.

\begin{table}[H]
\centering\footnotesize
\caption{Deviations after the registration (for the full-data model) and after the amendment (for the released model), with their effect on the reported results.}
\label{tab:deviations}
\begin{tabular}{>{\raggedright\arraybackslash}p{8.6cm}>{\raggedright\arraybackslash}p{6.4cm}}
\toprule
Deviation & Effect on the reported results \\
\midrule
After the registration, exploratory checks of the full-data model were added: accuracy by whether an annotated evidence span lies inside the input, the overlap of EI's articles with PubMedQA, duplicates across EI's splits, and intervals for the groups' differences. & None on any pre-registered number; the overlap check is in Section~\ref{sec:limits}. \\[2pt]
Gemma 4 31B-it's GPU time in the full-data model's evaluation was split between EI and PubMedQA by item count, which no plan stated. & Only the PubMedQA cost, which it overstates (Appendix~\ref{app:cost}). \\[2pt]
A top-level configuration file was added to the release folder so that Hugging Face counts downloads. & None. \\[2pt]
TabICLv2, a candidate member of the new readmission ensemble, was not trained, because of an error in the project's version check, not because it could not run; no plan stated this. & The new ensemble was fitted without it, and its pre-registered comparison with LightGBM was not computed; the selection rule then kept the earlier ensemble (Appendix~\ref{app:other}). \\[2pt]
The rented GPU installed vLLM \idn{0.30.0}'s default build for CUDA \idn{13.0} (NVIDIA's GPU computing platform), as the full-data model's plan allowed for a driver of CUDA \idn{13.0} or later (it reported \idn{13.2}), while the released model's plan had named the CUDA \idn{12.9} build among its fixed versions; no plan stated this change. & None: the pre-registered procedure and Gemma 4 31B-it's code, revision, prompt and reading of its answers were unchanged. \\[2pt]
BioGPT-Large-PubMedQA's checkpoint arguments could not be read (their host did not resolve). & It ran with the fallback settings fixed in advance; its accuracy, 0.7520, is below the 81.0 percent that \citet{luo2022} report, so it may be understated (Appendix~\ref{app:other}). \\[2pt]
The third run of the chosen configuration was left uncalibrated by an omission and calibrated afterward. & Its own figures are exploratory (Appendix~\ref{app:runs}); its weights are part of the released soup as planned. \\
\bottomrule
\end{tabular}
\end{table}

The registration was filed under an embargo until 1 October 2027; the amendment said that the embargo would end on the day of the release. The release was made public on 2 October 2026, and OSF showed the registration public when it was read at 01:06 UTC on 3 October 2026, which was still the evening before in New York.

\section{Per-class results and calibration}
\label{app:answers}

Table~\ref{tab:answers} gives the released model's F1 for each class and its calibration slope and intercept, each class against the rest (pre-registered secondary results). For each class the slope's interval includes 1 and the intercept's includes 0: no miscalibration of any class was detected. The slopes' intervals all lie within 0.7943 to 1.1833, and the intercepts' within $-$0.5151 to 0.5996.

\begin{table}[H]
\centering\footnotesize
\caption{The released model on Evidence Inference 2.0's test split, each class against the rest: F1, and the calibration slope (the coefficient of a logistic regression of the class's indicator on the logit of its calibrated probability; 1 is ideal) and intercept (with that logit as an offset; 0 is ideal), each with its 95 percent interval from 1,000 article resamples.}
\label{tab:answers}
\begin{tabular}{lcccr}
\toprule
Class & F1 & Calibration slope & Calibration intercept & Prompts \\
\midrule
Significantly increased & \begin{tabular}[t]{@{}c@{}}0.9185\\[-1pt]{\scriptsize (0.8940 to 0.9390)}\end{tabular} & \begin{tabular}[t]{@{}c@{}}0.9112\\[-1pt]{\scriptsize (0.8188 to 1.0405)}\end{tabular} & \begin{tabular}[t]{@{}c@{}}0.2244\\[-1pt]{\scriptsize ($-$0.1219 to 0.5996)}\end{tabular} & 411 \\[2pt]
Significantly decreased & \begin{tabular}[t]{@{}c@{}}0.9158\\[-1pt]{\scriptsize (0.8876 to 0.9437)}\end{tabular} & \begin{tabular}[t]{@{}c@{}}0.9919\\[-1pt]{\scriptsize (0.8773 to 1.1833)}\end{tabular} & \begin{tabular}[t]{@{}c@{}}$-$0.1294\\[-1pt]{\scriptsize ($-$0.4867 to 0.1917)}\end{tabular} & 294 \\[2pt]
No significant difference & \begin{tabular}[t]{@{}c@{}}0.9399\\[-1pt]{\scriptsize (0.9218 to 0.9555)}\end{tabular} & \begin{tabular}[t]{@{}c@{}}0.8850\\[-1pt]{\scriptsize (0.7943 to 1.0011)}\end{tabular} & \begin{tabular}[t]{@{}c@{}}$-$0.1139\\[-1pt]{\scriptsize ($-$0.5151 to 0.2811)}\end{tabular} & 513 \\[2pt]
\bottomrule
\end{tabular}

\end{table}

Table~\ref{tab:k1answers} gives, with the primary analysis, each class's F1: the released model minus Gemma 4 31B-it and minus the class-prior baseline (the other differences reported with it are in Tables~\ref{tab:bar} and~\ref{tab:main}). The class-prior baseline answers every prompt ``no significant difference'', so its F1 for the two directions is zero.

\begin{table}[H]
\centering\footnotesize
\caption{Per-class F1, reported with the primary analysis: the F1 of the released model on the 1,218 test prompts, and its difference from Gemma 4 31B-it's and from the class-prior baseline's, each with its 95 percent interval from 1,000 article resamples. \textsuperscript{\dag}No difference was detected: the interval includes zero.}
\label{tab:k1answers}
\begin{tabular}{lccc}
\toprule
Class & F1, released model & Minus Gemma 4 31B-it's & Minus the class prior's \\
\midrule
Significantly increased & \begin{tabular}[t]{@{}c@{}}0.9185\\[-1pt]{\scriptsize (0.8940 to 0.9390)}\end{tabular} & \begin{tabular}[t]{@{}c@{}}0.0712\\[-1pt]{\scriptsize (0.0366 to 0.1078)}\end{tabular} & \begin{tabular}[t]{@{}c@{}}0.9185\\[-1pt]{\scriptsize (0.8940 to 0.9390)}\end{tabular} \\[2pt]
Significantly decreased & \begin{tabular}[t]{@{}c@{}}0.9158\\[-1pt]{\scriptsize (0.8876 to 0.9437)}\end{tabular} & \begin{tabular}[t]{@{}c@{}}0.0858\\[-1pt]{\scriptsize (0.0488 to 0.1262)}\end{tabular} & \begin{tabular}[t]{@{}c@{}}0.9158\\[-1pt]{\scriptsize (0.8876 to 0.9437)}\end{tabular} \\[2pt]
No significant difference & \begin{tabular}[t]{@{}c@{}}0.9399\\[-1pt]{\scriptsize (0.9218 to 0.9555)}\end{tabular} & \begin{tabular}[t]{@{}c@{}}0.0170\textsuperscript{\dag}\\[-1pt]{\scriptsize ($-$0.0021 to 0.0366)}\end{tabular} & \begin{tabular}[t]{@{}c@{}}0.3472\\[-1pt]{\scriptsize (0.3123 to 0.3804)}\end{tabular} \\[2pt]
\bottomrule
\end{tabular}

\end{table}

\section{The cost per 1,000 questions}
\label{app:cost}

The cost per 1,000 questions is 1,000 times the GPU seconds per question, divided by 3,600 seconds an hour, times the rented GPU's hourly rate, \$3.52. For the released model the seconds are its forward passes over the test prompts; for Gemma 4 31B-it on EI they are its whole run's time on the GPU over its three passes (the first answers, the answers asked again with more log-probabilities, and the second process), model loading not counted; each job ran alone on the GPU (Table~\ref{tab:cost}). The PubMedQA model and Gemma 4 31B-it on PubMedQA were not run again on the test split for the released model, so their seconds come from the full-data model's evaluation, priced at the same hourly rate. In that evaluation, Gemma 4 31B-it's process answered EI and PubMedQA items together and its time was split between them by item count, which overstates PubMedQA's time: split by prompt tokens, it would be about 0.174 seconds a PubMedQA item (computed in the repository's report on the full-data model). The cost is not computed for readmission, whose trees run on the CPU.

\begin{table}[H]
\centering\footnotesize
\caption{The GPU time per question and the cost per 1,000 questions at the rented GPU's rate, \$3.52 an hour. GPU is graphics processing unit. Gemma 4 31B-it's PubMedQA time is its share by item count in the full-data model's evaluation, which overstates it (see the text).}
\label{tab:cost}
\begin{tabular}{llrr}
\toprule
Task & Model & \shortstack{GPU seconds\\ per question} & \shortstack{Dollars per\\ 1,000 questions} \\
\midrule
Evidence Inference & Released model & 0.3034 & \$0.297 \\[2pt]
Evidence Inference & Gemma 4 31B-it & 2.4982 & \$2.443 \\[2pt]
PubMedQA & PubMedQA model & 0.0200 & \$0.020 \\[2pt]
PubMedQA & Gemma 4 31B-it & 0.8040 & \$0.786 \\[2pt]
\bottomrule
\end{tabular}

\end{table}

\section{The PubMedQA and readmission models}
\label{app:other}

\paragraph{The release.} Fiorillo v0.5 holds one model for each task, and the release is judged task by task: the EI model of the main text, a PubMedQA model and a readmission model.

\paragraph{PubMedQA.} The PubMedQA model, from which the EI model is initialized, is Qwen3-4B-Base with the same kind of classification head, and its earlier calibration, unchanged in this release. It is evaluated on the 500 expert-labeled test items of PubMedQA \citep{jin2019}. It learned from PubMedQA's abstracts under PubMedQA's MIT License; unlike EI's articles, those abstracts' own licenses were not checked one by one, and most PubMed abstracts are the publishers'. Its labels for PubMedQA's unlabeled abstracts are pseudo-labels from an earlier model of mine built on Fiorillo reader v1 \citep{reader1}, an encoder further pretrained on PubMed abstracts with every PubMedQA article removed; no outside language model labeled them.

On PubMedQA's 500 test items, the PubMedQA model's accuracy was 0.7960 (0.7600 to 0.8340) and its log loss 0.5797 (Table~\ref{tab:pubmedqa}). For its class ``no'', the calibration slope was 1.1869 (1.0296 to 1.4260): the interval excludes 1, so miscalibration was detected, with probabilities less extreme than the outcomes support (Table~\ref{tab:pqaanswers}); for ``yes'' and ``maybe'' no miscalibration was detected. Against BioGPT-Large-PubMedQA, reported uncalibrated as its primary result, the model's log loss was lower by 0.3102 (0.1996 to 0.4219) and its accuracy higher by 0.0440 (0.0139 to 0.0780). BioGPT-Large-PubMedQA's accuracy here, 0.7520, is below the 81.0 percent that \citet{luo2022} report for it; its checkpoint arguments could not be read, so it ran with fallback settings fixed in advance (Appendix~\ref{app:deviations}), and it is a reference only. Against Gemma 4 31B-it, BioGPT-Large-PubMedQA's log loss was higher by 0.2768 (0.1695 to 0.3925); for accuracy no difference was detected, $-$0.0320 ($-$0.0680 to 0.0040). These test items were used in the evaluation of every version, so these results are not confirmatory.

\paragraph{Readmission.} The readmission task asks whether a hospital encounter of a patient with diabetes was followed by no readmission, a readmission after 30 days, or one within 30 days, on the University of California, Irvine (UCI) Diabetes 130-US hospitals data \citep[CC BY 4.0;][]{strack2014}; its test split holds 6,998 encounters. The readmission model is a weighted ensemble of gradient-boosted trees, XGBoost \citep{chen2016} with weight 0.4 and CatBoost \citep{prokhorenkova2018} with weight 0.6, with a temperature of 0.9957; a selection rule fixed in advance kept this ensemble, from the earlier version (Fiorillo v0.4), over a new one. On the 6,998 test encounters, its log loss was 0.8275 against the class-prior baseline's 0.8861, and its AUROC (area under the receiver operating characteristic curve) for readmission within 30 days was 0.6622 (Table~\ref{tab:readmission}). TabICLv2, a pre-registered candidate member of the new ensemble, was not trained because of an error in the project's version check, so the new ensemble was fitted without it and the pre-registered comparison of TabICLv2 with LightGBM \citep{ke2017} was not computed (Appendix~\ref{app:deviations}). The new ensemble's out-of-fold log loss, 0.8368, was not below the earlier ensemble's, 0.8372, by the 0.002 the rule required, so the release kept the earlier ensemble, and the difference between the ensembles is zero. For the XGBoost and CatBoost trees refitted with recoded diagnosis fields, neither of them in the released ensemble, no difference in log loss from their earlier counterparts was detected: 0.0016 ($-$0.0004 to 0.0037) and $-$0.0002 ($-$0.0023 to 0.0017). Log loss is the selection rule's measure. On other pre-registered metrics, the recoded XGBoost's Brier score was higher by 0.0014 (0.0000 to 0.0028) and its quadratic weighted kappa lower by 0.0129 (0.0006 to 0.0248), and the 30-day calibration intercept moved for both trees, by 0.0130 (0.0080 to 0.0184) and $-$0.0076 ($-$0.0126 to $-$0.0024). Each interval is a 95 percent interval from 1,000 item resamples of the 6,998 test encounters. This test split, too, was used in the evaluation of every version.

\begin{table}[H]
\centering\footnotesize
\caption{PubMedQA, the 500 expert-labeled test items (this test split was used in the evaluation of every version, so these results are not confirmatory). Each cell gives the estimate and its 95 percent interval from 1,000 item resamples; ECE is the expected calibration error. BioGPT-Large-PubMedQA was fine-tuned on sets that hold all 150 validation items, so its uncalibrated probabilities are its primary result and its calibration on those items is secondary.}
\label{tab:pubmedqa}
\begin{tabular}{>{\raggedright\arraybackslash}p{3.0cm}ccccc}
\toprule
System & Accuracy & Macro-F1 & Log loss & Brier score & ECE \\
\midrule
PubMedQA model & \begin{tabular}[t]{@{}c@{}}0.7960\\[-1pt]{\scriptsize (0.7600 to 0.8340)}\end{tabular} & \begin{tabular}[t]{@{}c@{}}0.5698\\[-1pt]{\scriptsize (0.5406 to 0.6039)}\end{tabular} & \begin{tabular}[t]{@{}c@{}}0.5797\\[-1pt]{\scriptsize (0.5070 to 0.6509)}\end{tabular} & \begin{tabular}[t]{@{}c@{}}0.3147\\[-1pt]{\scriptsize (0.2683 to 0.3595)}\end{tabular} & \begin{tabular}[t]{@{}c@{}}0.0401\\[-1pt]{\scriptsize (0.0315 to 0.0800)}\end{tabular} \\[2pt]
Gemma 4 31B-it & \begin{tabular}[t]{@{}c@{}}0.7840\\[-1pt]{\scriptsize (0.7460 to 0.8200)}\end{tabular} & \begin{tabular}[t]{@{}c@{}}0.5862\\[-1pt]{\scriptsize (0.5527 to 0.6252)}\end{tabular} & \begin{tabular}[t]{@{}c@{}}0.6131\\[-1pt]{\scriptsize (0.5309 to 0.6957)}\end{tabular} & \begin{tabular}[t]{@{}c@{}}0.3209\\[-1pt]{\scriptsize (0.2728 to 0.3696)}\end{tabular} & \begin{tabular}[t]{@{}c@{}}0.0575\\[-1pt]{\scriptsize (0.0431 to 0.0970)}\end{tabular} \\[2pt]
BioGPT-Large-PubMedQA, uncalibrated & \begin{tabular}[t]{@{}c@{}}0.7520\\[-1pt]{\scriptsize (0.7120 to 0.7900)}\end{tabular} & \begin{tabular}[t]{@{}c@{}}0.5481\\[-1pt]{\scriptsize (0.5119 to 0.5922)}\end{tabular} & \begin{tabular}[t]{@{}c@{}}0.8899\\[-1pt]{\scriptsize (0.7428 to 1.0347)}\end{tabular} & \begin{tabular}[t]{@{}c@{}}0.4058\\[-1pt]{\scriptsize (0.3462 to 0.4623)}\end{tabular} & \begin{tabular}[t]{@{}c@{}}0.1371\\[-1pt]{\scriptsize (0.1114 to 0.1803)}\end{tabular} \\[2pt]
BioGPT-Large-PubMedQA, calibrated (secondary) & \begin{tabular}[t]{@{}c@{}}0.7380\\[-1pt]{\scriptsize (0.6980 to 0.7761)}\end{tabular} & \begin{tabular}[t]{@{}c@{}}0.5269\\[-1pt]{\scriptsize (0.4963 to 0.5638)}\end{tabular} & \begin{tabular}[t]{@{}c@{}}0.7340\\[-1pt]{\scriptsize (0.6341 to 0.8367)}\end{tabular} & \begin{tabular}[t]{@{}c@{}}0.3886\\[-1pt]{\scriptsize (0.3352 to 0.4389)}\end{tabular} & \begin{tabular}[t]{@{}c@{}}0.0876\\[-1pt]{\scriptsize (0.0667 to 0.1299)}\end{tabular} \\[2pt]
Class prior & \begin{tabular}[t]{@{}c@{}}0.5520\\[-1pt]{\scriptsize (0.5080 to 0.5940)}\end{tabular} & \begin{tabular}[t]{@{}c@{}}0.2371\\[-1pt]{\scriptsize (0.2246 to 0.2484)}\end{tabular} & \begin{tabular}[t]{@{}c@{}}1.1593\\[-1pt]{\scriptsize (1.0401 to 1.2920)}\end{tabular} & \begin{tabular}[t]{@{}c@{}}0.5855\\[-1pt]{\scriptsize (0.5503 to 0.6208)}\end{tabular} & \begin{tabular}[t]{@{}c@{}}0.0646\\[-1pt]{\scriptsize (0.0226 to 0.1086)}\end{tabular} \\[2pt]
\bottomrule
\end{tabular}

\end{table}

\begin{table}[H]
\centering\footnotesize
\caption{The PubMedQA model on PubMedQA's 500 test items, each class against the rest: the calibration slope (1 is ideal) and intercept (0 is ideal), defined as in Table~\ref{tab:answers}, each with its 95 percent interval from 1,000 item resamples (this test split was used in every version, so these results are not confirmatory).}
\label{tab:pqaanswers}
\begin{tabular}{lccr}
\toprule
Class & Calibration slope & Calibration intercept & Items \\
\midrule
Yes & \begin{tabular}[t]{@{}c@{}}0.9647\\[-1pt]{\scriptsize (0.8394 to 1.1381)}\end{tabular} & \begin{tabular}[t]{@{}c@{}}$-$0.0217\\[-1pt]{\scriptsize ($-$0.2800 to 0.2453)}\end{tabular} & 276 \\[2pt]
No & \begin{tabular}[t]{@{}c@{}}1.1869\\[-1pt]{\scriptsize (1.0296 to 1.4260)}\end{tabular} & \begin{tabular}[t]{@{}c@{}}$-$0.1513\\[-1pt]{\scriptsize ($-$0.3927 to 0.0973)}\end{tabular} & 169 \\[2pt]
Maybe & \begin{tabular}[t]{@{}c@{}}0.8990\\[-1pt]{\scriptsize (0.5392 to 1.3122)}\end{tabular} & \begin{tabular}[t]{@{}c@{}}0.2371\\[-1pt]{\scriptsize ($-$0.0890 to 0.5250)}\end{tabular} & 55 \\[2pt]
\bottomrule
\end{tabular}

\end{table}

\begin{table}[H]
\centering\footnotesize
\caption{Readmission on the University of California, Irvine, Diabetes 130-US hospitals data, the 6,998 test encounters (this test split was used in the evaluation of every version, so these results are not confirmatory). Each cell gives the estimate and its 95 percent interval from 1,000 item resamples. ECE is the expected calibration error, whose percentile interval can lie above its estimate, since the ECE is biased upward in resamples; AUROC is the area under the receiver operating characteristic curve; it and the calibration intercept and slope are for readmission within 30 days against the rest. The calibration slope is not computed for the class prior, whose prediction is the same for every encounter.}
\label{tab:readmission}
\setlength{\tabcolsep}{3.5pt}
\begin{tabular}{>{\raggedright\arraybackslash}p{2.0cm}cccccc}
\toprule
System & Log loss & ECE & Accuracy & \shortstack{AUROC,\\ 30 days} & \shortstack{Calibration\\ intercept, 30 days} & \shortstack{Calibration\\ slope, 30 days} \\
\midrule
Tree ensemble & \begin{tabular}[t]{@{}c@{}}0.8275\\[-1pt]{\scriptsize (0.8132 to 0.8411)}\end{tabular} & \begin{tabular}[t]{@{}c@{}}0.0066\\[-1pt]{\scriptsize (0.0085 to 0.0225)}\end{tabular} & \begin{tabular}[t]{@{}c@{}}0.6157\\[-1pt]{\scriptsize (0.6042 to 0.6269)}\end{tabular} & \begin{tabular}[t]{@{}c@{}}0.6622\\[-1pt]{\scriptsize (0.6394 to 0.6855)}\end{tabular} & \begin{tabular}[t]{@{}c@{}}$-$0.0192\\[-1pt]{\scriptsize ($-$0.1031 to 0.0646)}\end{tabular} & \begin{tabular}[t]{@{}c@{}}1.0493\\[-1pt]{\scriptsize (0.9108 to 1.1940)}\end{tabular} \\[2pt]
Class prior & \begin{tabular}[t]{@{}c@{}}0.8861\\[-1pt]{\scriptsize (0.8734 to 0.8984)}\end{tabular} & \begin{tabular}[t]{@{}c@{}}0.0017\\[-1pt]{\scriptsize (0.0002 to 0.0127)}\end{tabular} & \begin{tabular}[t]{@{}c@{}}0.5942\\[-1pt]{\scriptsize (0.5829 to 0.6047)}\end{tabular} & \begin{tabular}[t]{@{}c@{}}0.5000\\[-1pt]{\scriptsize (0.5000 to 0.5000)}\end{tabular} & \begin{tabular}[t]{@{}c@{}}$-$0.0299\\[-1pt]{\scriptsize ($-$0.1120 to 0.0536)}\end{tabular} & not computed \\[2pt]
\bottomrule
\end{tabular}

\end{table}

\end{document}